\pdfoutput=1
\documentclass[10pt,twocolumn,a4paper]{article}
\usepackage[a4paper,margin=1.8cm,columnsep=0.6cm]{geometry}
\usepackage[T1]{fontenc}
\usepackage{mathptmx}
\usepackage{amsmath,amssymb}
\usepackage{booktabs}
\usepackage{graphicx}
\usepackage{xcolor}
\usepackage{cite}
\usepackage[hidelinks]{hyperref}
\usepackage{caption}
\title{\Large\bfseries Seen, Exists, Valid: Calibrated Validity of Stored 3D Geometry\\ Across Long Observation Gaps}
\author{Anwaar Ulhaq\\ \small School of Engineering and Technology, Central Queensland University, Sydney, Australia\\ \small\texttt{a.anwaarulhaq@cqu.edu.au}}
\date{}

\begin{document}
\makeatletter
\twocolumn[{%
\@maketitle
\vspace{-0.6em}
\begin{center}
\includegraphics[width=\textwidth]{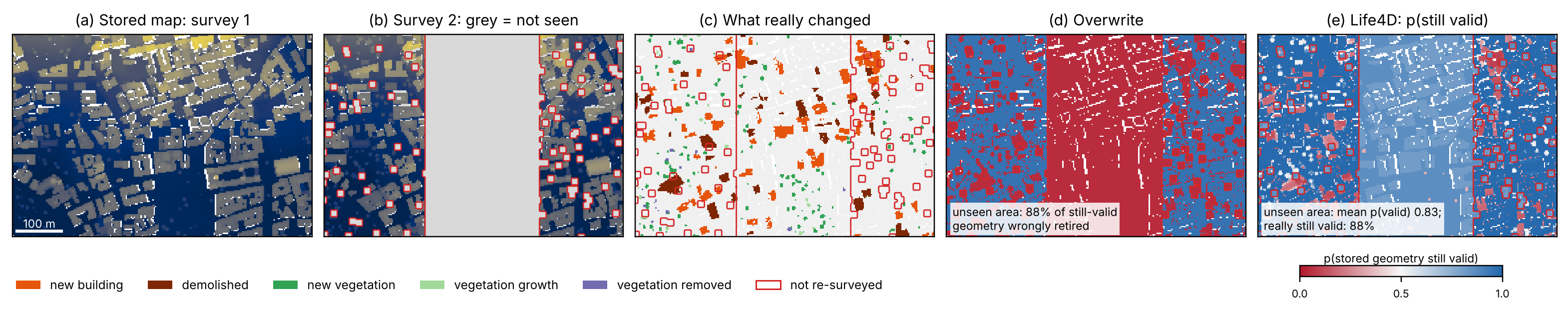}
\captionof{figure}{The problem on real benchmark data (Urb3DCD-v2, LyonS test scene, 3\,m cells). (a)~A stored 3D map from survey~1. (b)~Survey~2 misses a flight strip and some occluded blocks (red outline). (c)~What really changed between the surveys, including inside the unseen strip. (d)~Overwriting the map treats ``not seen'' as ``gone'' and retires 88\% of the unseen geometry that is in fact still valid. Keeping the last observation (not shown) instead marks all unseen geometry as valid with confidence 0.95, although 12\% of it changed. (e)~Life4D keeps observation separate from state: observed cells follow the new evidence, while unseen cells keep their geometry with a class-conditioned confidence learned from other scenes (mean 0.83, against 88\% truly still valid). Changed unseen cells are not yet identified individually; that needs change rates learned from repeated revisits (Section~\ref{sec:lim}).}\label{fig:teaser}
\end{center}
\vspace{0.4em}
}]
\makeatother

\begin{abstract}
Repeated airborne LiDAR surveys are costly, while stored 3D geometry can become outdated between observations. Existing change-detection methods compare observed epochs. However, they do not quantify the validity of geometry that has not been revisited. We distinguish observation, existence and geometric validity using a probabilistic filter over five geometric transitions: persistence, modification, growth, removal and replacement. The framework learns class-specific continuous-time transition priors from interval-censored observations and combines shape residuals with LiDAR free-space evidence. Functional anchors enable prior sharing across sites with different survey frequencies. In controlled experiments with misspecified dynamics and 60\% visibility, our method reduces combined error from 36.9\% to 16.5\% and expected calibration error (ECE) from 0.19--0.32 to 0.017. Cross-site sharing reduces ECE at two-survey sites from 0.156 to 0.029. On Urb3DCD-v2, the method achieves 9.8\% combined error and 0.016 ECE, outperforming baselines at 12.6\% and at least 0.070, respectively. Re-survey prioritisation improves only modestly over random selection with two observations. Code and an evaluation protocol are released at \url{https://github.com/anwaarua/life4d}.
\end{abstract}

\section{Introduction}
Governments and industry now capture the built and natural environment in 3D repeatedly. Between captures, buildings are modified, vegetation grows, mine faces advance and assets are removed. A user of a stored 3D model therefore needs to know which parts of it are still true, and with what confidence, especially in areas that were not re-surveyed. Overwriting the model with each new survey treats everything unseen as gone. Keeping the last observation treats everything unseen as unchanged. Neither is calibrated. Figure~\ref{fig:teaser} shows this on a benchmark city scene: a second survey misses a strip, overwriting discards almost all of the still-valid geometry there, and keeping the last observation is confidently wrong about the 12\% that changed. What a user needs instead is a calibrated answer to a different question: not ``what did the new survey see?'' but ``how far can I still trust each part of the stored map?''

We make four contributions. (i)~A formulation of \emph{geometric validity} that keeps observation status separate from physical state and from historical truth. (ii)~A multi-state filter with a class-conditioned, continuous-time transition prior learned from interval-censored revisits, and a free-space observation rule under which non-observation never implies removal. (iii)~Cross-site pooling of transition priors through functional anchors on like structures. (iv)~An evaluation protocol with stale-geometry error, false retirement, calibration and re-survey efficiency, reported on a controlled study and on bi-temporal airborne LiDAR.

\section{Related work}
\textbf{LiDAR change detection.} Multi-temporal 3D change detection compares airborne LiDAR or photogrammetric epochs \cite{stilla}. Deep Siamese networks label several change types directly from bi-temporal point clouds \cite{skpconv}, evaluated on simulated (Urb3DCD) and national (AHN-CD) surveys \cite{ahncd}. These methods compare two epochs; they maintain no identity, history or uncertainty for unobserved areas.
\textbf{Point-cloud representations.} Point Transformer V3 and Sonata provide general point features \cite{ptv3,sonata}; hyperspectral--LiDAR fusion adds material evidence \cite{lgcf}. They describe a single time.
\textbf{Change mapping and persistence in robotics.} Object-level volumetric maps detect change on revisit \cite{pmtsdf,pocd,khronos}; Gaussian maps update scenes that change out of view \cite{game}; Living Scenes relocalises objects across rescans but treats unmatched instances as removed \cite{living}. LT-Mem keeps object identity, event history and a volatility score from monocular video \cite{ltmem}. Persistence filters model the survival of semi-static landmarks \cite{rosen,fremen,perpetua}, and lifelong graph SLAM bounds memory by information rather than time \cite{kretzschmar}. These address binary presence of landmarks or objects, mostly from one sensor, without dense multi-state geometric validity or calibration.

\section{Problem formulation}
A stored map holds entities $i$ (objects or surface cells) with geometry $G_i$, identity and dated versions. At time $t$ each entity has a physical state $Z^i_t\in\mathcal{Z}=\{\text{persistent, modified, grown, removed, replaced}\}$ and an observation status $O^i_t\in\{\text{seen, unseen, occluded}\}$. Stored geometry is \emph{valid} at $t$ if $Z^i_t=\text{persistent}$ since its last version. The task is to output a calibrated $p(\text{valid}^i_t\mid\mathcal{D}_{1:t})$ for every entity, including those with $O^i_t=\text{unseen}$.
\textbf{Metrics.} With threshold 0.5: \emph{stale} = fraction of queries where changed geometry is treated as valid; \emph{false retirement} = fraction where still-valid geometry is treated as invalid; their sum; expected calibration error (ECE, 10 bins) \cite{guo}; and, for unobserved entities, \emph{re-survey efficiency}: the share of truly changed entities captured when only the 20\% highest-risk entities are re-surveyed (random choice captures 20\%).

\section{Method}
\textbf{Filter.} Validity is the filtered posterior
\begin{multline}
p(Z^i_t\mid\mathcal{D}_{1:t})\propto p(o^i_t\mid Z^i_t)\\
\times\sum_{z'}p(Z^i_t\mid Z^i_{t-1}{=}z',\Delta t,c^i)\,p(Z^i_{t-1}{=}z'\mid\mathcal{D}_{1:t-1}),
\end{multline}
where non-persistent states are absorbing until a new version is created.
\textbf{Transition prior.} The probability of leaving the persistent state within $\Delta t$ is $1-\exp(-\lambda(c)\Delta t)$, with $c$ the class and context. Without observation, confidence in validity can only decrease.
\textbf{Learning from censored revisits.} A revisit pair where the entity was seen unchanged at $L_i$ and first seen changed at $R_i$ (or never, $R_i=\infty$) contributes $\exp(-\lambda L_i)-\exp(-\lambda R_i)$ to the likelihood, so every pair is used, including the majority with no change. We maximise this per class.
\textbf{Observation model.} The likelihood combines the geometric residual after alignment, spectral or appearance distance when available, and free-space evidence: a laser ray passing through stored geometry supports removal, whereas absence of rays supports only ``unseen''.
\textbf{Decision rule.} Decay moves probability towards ``possibly changed'', never directly to ``removed''; geometry is retired only with positive removal evidence and free-space support, and otherwise is flagged for re-observation.
\textbf{Cross-site pooling.} Like structures (e.g.\ haul roads: crest, centreline, toe, drain) are registered to functional anchors, as homologous landmarks are in morphometrics \cite{bookstein,gower}; sites of the same anchored class share $\lambda(c)$, so a two-survey site uses a prior learned across sites rather than from its own few revisits.

\section{Experiments}
\subsection{Controlled study}
\begin{figure}[t]
\centering
\includegraphics[width=\columnwidth]{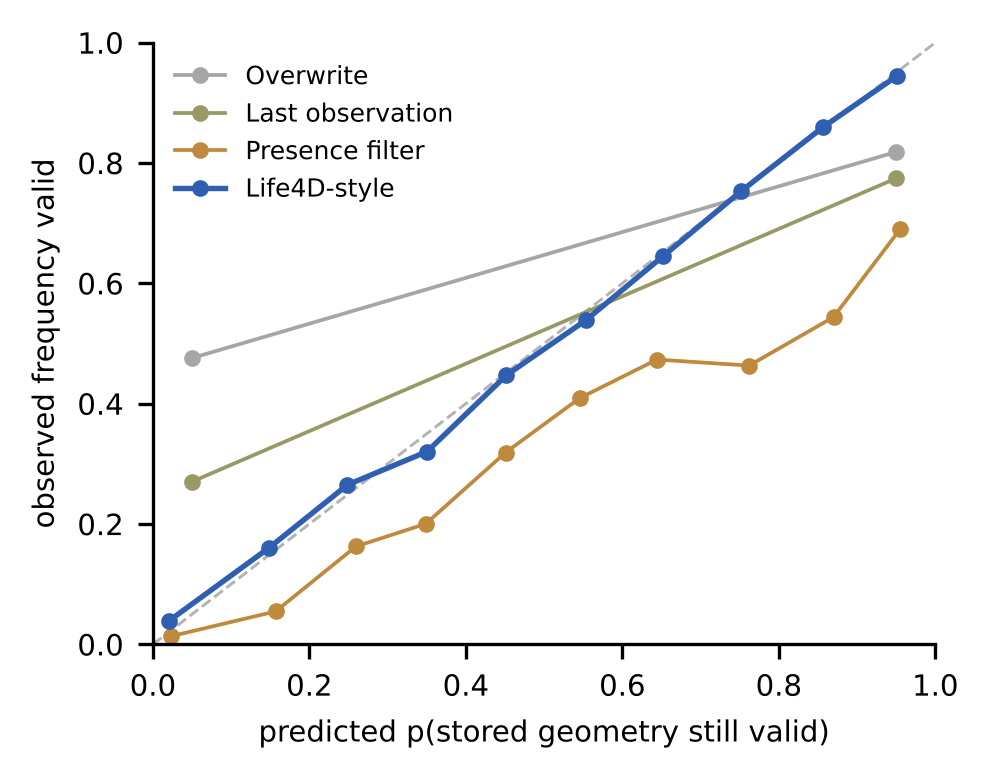}
\caption{Reliability of predicted validity (controlled study, all seeds). Only the validity filter follows the diagonal.}
\label{fig:rel}
\end{figure}

8,000 entities in 200 scenes of four classes (buildings, roads, vegetation, small objects) change at Weibull-distributed times, deliberately misspecifying the exponential prior. Each entity is seen at 60\% of irregular revisits 2--14 months apart, with noisy presence and residual evidence; validity is queried 12--60 months after first capture on held-out scenes (half the scenes for training), 10 seeds. Table~\ref{tab:sim} and Figure~\ref{fig:rel} show the results. Overwriting retires almost a third of still-valid geometry because it equates non-observation with removal. Keeping the last observation never decays and is overconfident. A presence-only filter misses modified and grown geometry. The validity filter has the lowest combined error and is close to calibrated, despite the misspecified dynamics.

\begin{table}[t]
\centering\small
\caption{Controlled study (held-out scenes) and Urb3DCD-v2 airborne LiDAR benchmark (held-out areas); mean $\pm$ s.d., 10 seeds.}
\label{tab:sim}
\setlength{\tabcolsep}{3pt}
\begin{tabular}{@{}lcccc@{}}
\toprule
Method & Stale\% & False ret.\% & Comb.\% & ECE\\
\midrule
Overwrite & 6.6$\pm$0.4 & 30.3$\pm$0.6 & 36.9 & 0.319\\
Last observation & 14.8$\pm$0.4 & 9.3$\pm$0.5 & 24.1 & 0.191\\
Presence filter \cite{rosen} & 29.9$\pm$0.6 & 1.4$\pm$0.1 & 31.4 & 0.247\\
\textbf{Validity filter (ours)} & \textbf{11.0$\pm$0.6} & \textbf{5.5$\pm$0.4} & \textbf{16.5} & \textbf{0.017}\\
\midrule
\multicolumn{5}{@{}l}{\emph{Urb3DCD-v2 airborne LiDAR (held-out areas)}}\\
Overwrite & 0.4$\pm$0.1 & 37.6$\pm$4.5 & 38.0 & 0.370\\
Last observation & 6.1$\pm$1.3 & 6.5$\pm$0.6 & 12.6 & 0.076\\
Presence filter \cite{rosen} & 15.3$\pm$0.7 & 0.0$\pm$0.0 & 15.3 & 0.070\\
\textbf{Validity filter (ours)} & 7.1$\pm$1.2 & 2.7$\pm$0.3 & \textbf{9.8} & \textbf{0.016}\\
\bottomrule
\end{tabular}
\end{table}

\subsection{Cross-site pooling}
We add 60 sites observed at only two surveys. Learning the change rate from each site's own revisits gives stale 30.2\%, false retirement 7.0\% and ECE 0.156. Using the rate pooled across anchored sites of the same class gives stale 11.9\%, false retirement 9.1\% and ECE 0.029 (10 seeds).

\subsection{Airborne LiDAR benchmark (Urb3DCD-v2)}
Urb3DCD-v2 \cite{urb3dcd} simulates airborne LiDAR over a real 3D city model of Lyon, with labelled changes between two epochs (new and demolished buildings; planted, grown and removed vegetation; mobile objects). We use its three test scenes (Lidar05 subset, 0.30--0.31 million points per epoch), which share one area with independent changes. Both epochs are rasterised to 3\,m surface cells (the point density leaves most 1\,m cells empty). Non-observation in the second epoch is simulated by removing one flight strip (25\% of the width) and random occlusion blocks (10\%). The area is divided into $4\times4$ spatial blocks; priors and likelihood are learned on half of the blocks and evaluated on the other half, so no test location is seen in training in any scene (10 random splits). Cell class is the stored semantic class of the first epoch; mobile objects are treated as transient; cells with no second-epoch return carry no label and are excluded.

The validity filter has the lowest combined error (9.8\% against 12.6\% for keeping the last observation and 38.0\% for overwriting) and the best calibration (ECE 0.016 against 0.070--0.370; Table~\ref{tab:sim}). Overwriting wrongly retires 85\% of still-valid unobserved geometry. On unobserved cells, keeping the last observation is poorly calibrated (ECE 0.104), while the filter (0.016) matches a presence filter with a global change rate (0.015). Ranking unobserved cells for re-survey improves only modestly on random choice: the top 20\% of cells by risk contain 27\% of the changes (AUROC 0.55). With two epochs, the class-conditioned prior adds little beyond the global rate, and the change rate of building cells varied seven-fold between the three scenes (4.7--33.9\%), which is the motivation for learning rates from repeated revisits and pooling them across anchored sites. The ordering of combined error and ECE was unchanged with 2\,m or 4\,m cells and with a label-free height-and-roughness class in place of stored semantics (combined 9.2--11.4\%, ECE 0.020--0.021).

\section{Limitations}\label{sec:lim}
The controlled study uses a known sensor model. Urb3DCD-v2 is simulated LiDAR over a real city model, and has two epochs, so it tests validity across one gap and re-survey ranking, not multi-revisit decay; its three test scenes share one area. Testing on real repeat surveys (AHN3/AHN4) is the next step. Cells are 2.5D; full 3D entity identity is future work.

\section{Conclusion}
Treating validity, not just change, as the target turns repeated surveys into maintained knowledge with calibrated confidence and a principled re-survey priority.

\noindent\textbf{Code.} Simulation, pilot code and evaluation protocol: \url{https://github.com/anwaarua/life4d}.

\noindent\textbf{Acknowledgements.} This work received no specific grant funding. Text drafting and code were assisted by an AI system (Claude, Anthropic). The author directed the work, ran and verified all code and results, and takes full responsibility for the content.

\end{document}